\documentclass[11pt,a4paper]{article}
\usepackage[hyperref]{eacl2021}
\usepackage{times}
\usepackage{amssymb}
\usepackage{amsmath}
\usepackage{subcaption}
\usepackage{latexsym}
\usepackage[pdftex]{graphicx}
\usepackage{tabularx}

\usepackage{collcell}

\definecolor{lightblue}{rgb}{0.19, 0.55, 0.91}
\definecolor{darkyellow}{rgb}{1.0, 0.65, 0.0}

\usepackage{microtype}

\aclfinalcopy 
\def\aclpaperid{7} 

\title{Synthetic Data Generation for Grammatical Error Correction with Tagged Corruption Models}

\author{Felix Stahlberg and Shankar Kumar \\
Google Research \\
\texttt{\{fstahlberg,shankarkumar\}@google.com}}

\date{}

\begin{document}
\maketitle
\begin{abstract}
Synthetic data generation is widely known to boost the accuracy of neural grammatical error correction (GEC) systems, but existing methods often lack diversity or are too simplistic to generate the broad range of grammatical errors made by human writers. In this work, we use error type tags from automatic annotation tools such as ERRANT to guide synthetic data generation. We compare several models that can produce an ungrammatical sentence given a clean sentence and an error type tag. We use these models to build a new, large synthetic pre-training data set with error tag frequency distributions matching a given development set. Our synthetic data set yields large and consistent gains, improving the state-of-the-art on the BEA-19 and CoNLL-14 test sets. We also show that our approach is particularly effective in adapting a GEC system, trained on mixed native and non-native English, to a native English test set, even surpassing real training data consisting of high-quality sentence pairs.
\end{abstract}

\section{Introduction}

Grammatical error correction (GEC) systems aim to automatically correct grammatical and other types of writing errors in text. It is common to view this problem as a sequence-to-sequence task (i.e.\ ungrammatical sentence $\rightarrow$ grammatical sentence) and borrow models that were originally developed for neural machine translation (NMT)~\cite{chollampatt2018multilayer, junczys-dowmunt-etal-2018-approaching, ge-fluency}.
Back-translation \citep{sennrich-etal-2016-improving} is a synthetic data generation technique for NMT that employs a translation system trained in the reverse direction to synthesize source sentences from sentences in the target language, and is still one of the most effective strategies to use monolingual data in NMT training. Similarly, synthetic training data generation for GEC has also been widely studied in the literature \citep{brockett-etal-2006-correcting,foster-etal-2009-generate,rozovskaya-roth-2010-generating,felice-etal-2014-grammatical,rei-etal-2017-artificial,kasewa-etal-2018-wronging,xie-etal-2018-noising,ge-etal-2018-fluency,ge-fluency,gec-pseudo-data,lichtarge-etal-2019-corpora,stahlberg-byrne-2019-cueds,zhao-etal-2019-improving,xu-etal-2019-erroneous,grundkiewicz-etal-2019-neural,choe-etal-2019-neural,takahashi-etal-2020-grammatical}.
This work is inspired by previous efforts to use ideas from back-translation for GEC \citep{kasewa-etal-2018-wronging,xie-etal-2018-noising,gec-pseudo-data}. In contrast to prior work, we use error type tags such as \texttt{SPELL} (spelling error) or \texttt{SVA} (subject-verb agreement error) to control the output of our corruption models and generate more realistic as well as diverse grammatical errors. Our tagged corruption models are trained to output the corrupted sentence given a clean sentence and an error tag, e.g.:
\begin{quote}
``\texttt{NOUN:INFL} There were a lot of sheep.'' $\rightarrow$ ``There were a lot of sheeps.''
\end{quote}
The tags mitigate the tendency of untagged corruption models to produce simplistic corruptions since many error type tags require more complex rewrites. In general, there is a one-to-many mapping from a clean sentence to a noisy sentence. Using a regular corruption model, many of these synthetic errors tend to be simplistic,\footnote{Example outputs from untagged and tagged corruption models can be found in Appendix \ref{sec:example-corruptions}.} but adding tag information allows the model to generate specific patterns of errors that can be found in actual GEC corpora. The benefit of covering a wide range of error types when generating pseudo data for GEC has also been demonstrated by \citet{takahashi-etal-2020-grammatical,wan-etal-2020-improving}.
Moreover, the tag distribution in the synthetic data can be made to match the distribution of a specific target domain. We use this distribution matching technique to adapt a GEC system to better correct errors by native speakers.

\begin{figure*}[t!]
  \begin{subfigure}[b]{0.33\textwidth}
    \centering
    \includegraphics[scale=0.13]{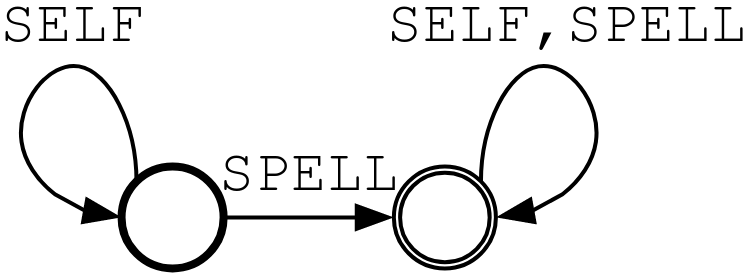}
    \caption{\sc{NoSigma}}
    \label{fig:no-sigma}
  \end{subfigure}
  \begin{subfigure}[b]{0.33\textwidth}
    \centering
    \includegraphics[scale=0.13]{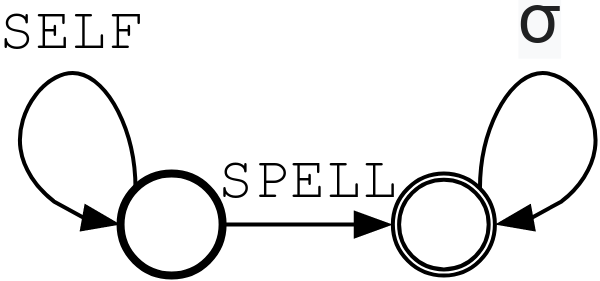}
    \caption{\sc{PostSigma}}
    \label{fig:post-sigma}
  \end{subfigure}
  \begin{subfigure}[b]{0.33\textwidth}
    \centering
    \includegraphics[scale=0.13]{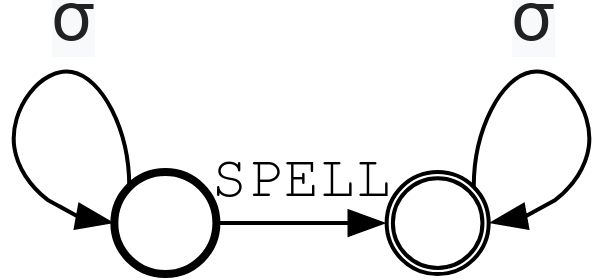}
    \caption{\sc{PrePostSigma}}
    \label{fig:pre-post-sigma}
  \end{subfigure}
  \caption{Constraining FSTs for the \texttt{SPELL} tag. The $\sigma$-self loops can match any tag. \texttt{SELF} is used by Seq2Edits to mark source spans that are not modified.}
  \label{fig:constraint-fsts}
\end{figure*}

As an alternative to adding the tag directly to the input sequence, we add inference-time constraints to the recently proposed Seq2Edits model~\citep{seq2edits} to force the generation of a particular tag. We implement such constraints using Finite State Transducers (FSTs). We then use these corruption models to generate synthetic training data that follows a desired tag distribution, for example the tag distribution on the development set. Using our new synthetic pre-training sets\footnote{\href{https://github.com/google-research-datasets/C4_200M-synthetic-dataset-for-grammatical-error-correction}{https://github.com/google-research-datasets/C4\_200M-synthetic-dataset-for-grammatical-error-correction}} we report state-of-the-art results on two popular GEC test sets (BEA-test: 74.9 F$_{0.5}$, CoNLL-14: 68.3 F$_{0.5}$). In our experiments on GEC for native English, a model fine-tuned on synthetic data that follows a native English error-tag distribution can even surpass a model fine-tuned on high-quality, real (i.e. non-synthetic) data.

\begin{table}[t!]
\small
\centering
\begin{tabularx}{\linewidth}{l@{\hspace{-0.6em}}c@{\hspace{0.05em}}X}
\hline
\textbf{Tag} & \textbf{Log-prob.} & \textbf{Corruption model output} \\
\hline
\texttt{ADJ} & \colorbox{yellow!14.61691454!red}{-3.06}	& There were a lot of many sheep.\\
\texttt{ADJ:FORM} & \colorbox{yellow!47.58839578!red}{-2.49}	& There were a more better of sheep.\\
\texttt{ADV} & \colorbox{yellow!39.84061712!red}{-2.63} & There were a lot of sheep there.\\
\texttt{CONJ} & \colorbox{green!17.21536149!yellow}{-1.39}	&	And there were a lot of sheep.\\
\texttt{CONTR} & \colorbox{green!59.77150773!yellow}{-0.90}	&	There're a lot of sheep.\\
\texttt{DET} & \colorbox{green!45.71760353!yellow}{-1.06}	&	There were lot of sheep.\\
\texttt{K} & \colorbox{green!11.91176357!yellow}{-1.45}	&	There were a lot of.\\
\texttt{MORPH} & \colorbox{green!79.83528688!yellow}{-0.67}	&	There were a lot of sheeps.\\
\texttt{NOUN} & \colorbox{yellow!0.0!red}{-3.31} &	There were a lot of seep.\\
\texttt{NOUN:INFL} & \colorbox{green!85.39430601!yellow}{-0.61}	&	There were a lot of sheeps.\\
\texttt{NOUN:NUM} & \colorbox{green!22.95729629!yellow}{-1.33}	&	There were a lots of sheep.\\
\texttt{NOUN:POSS} & \colorbox{green!58.62341636!yellow}{-0.92}	&	There were a lot of sheep's.\\
\texttt{ORTH} & \colorbox{green!69.90933017!yellow}{-0.79}	&	There were alot of sheep.\\
\texttt{OTHER} & \colorbox{yellow!41.29487641!red}{-2.60}	& There were many sheep.\\
\texttt{PART} & \colorbox{green!32.46221661!yellow}{-1.22}	&	There were a lot off sheep.\\
\texttt{PREP} & \colorbox{green!44.06988463!yellow}{-1.08}	&	There were a lot sheep.\\
\texttt{PRON} & \colorbox{green!3.248133582!yellow}{-1.55}	&	It was a lot of sheep.\\
\texttt{PUNCT} & \colorbox{green!51.51230915!yellow}{-1.00}	&	There were a lot of sheep\\
\texttt{SPELL} & \colorbox{yellow!30.46425359!red}{-2.79}	& There were a lot of sheeps.\\
\texttt{VERB} & \colorbox{yellow!42.57967984!red}{-2.58}	& There had a lot of sheep.\\
\texttt{VERB:FORM} & \colorbox{yellow!56.48874085!red}{-2.34} &	There being a lot of sheep.\\
\texttt{VERB:INFL} & \colorbox{green!43.51027051!yellow}{-1.09}	&	There were a lot of sheeps.\\
\texttt{VERB:SVA} & \colorbox{green!100.0!yellow}{-0.44} &	There was a lot of sheep.\\
\texttt{VERB:TENSE} & \colorbox{green!88.82893011!yellow}{-0.57}	&	There are a lot of sheep.\\
\texttt{WO} & \colorbox{yellow!83.67234551!red}{-1.87} & There were a lot sheep of.\\
\hline
\end{tabularx}
\caption{\label{tab:sheep-corruptions} Outputs of a tagged Seq2Edits corruption model for the example input sentence ``There were a lot of sheep.''. The ERRANT error type tags are described in \citet{bryant-etal-2017-automatic}.}
\end{table}

\section{Tagged Corruption Models}
\label{sec:tagged-corruption}

At the core of our approach is a model that generates an ungrammatical sentence from a clean sentence given an error tag $t\in \mathcal{T}$ that describes the desired type of error. $\mathcal{T}$ is the set of 25 error type tags supported by the automatic annotation toolkit ERRANT \citep{felice-etal-2016-automatic,bryant-etal-2017-automatic}.

A straightforward way to train such a tagged corruption model is to annotate a parallel corpus with ERRANT, prepend the ERRANT tag to the clean sentence, and train a model such as a standard Transformer \citep{transformer} to generate the ungrammatical sentence.\footnote{If a sentence pair has multiple tags we duplicate it in the training set for each unique tag. This potentially enables the corruption model to learn co-occurrences of error categories since multiple errors may be labelled with a single tag.} This idea is similar to the multi-lingual NMT system of \citet{johnson-etal-2017-googles} that adds the target language ID tag to the source sentence.

\begin{figure*}[t!]
\centering
\small
\includegraphics[scale=0.19]{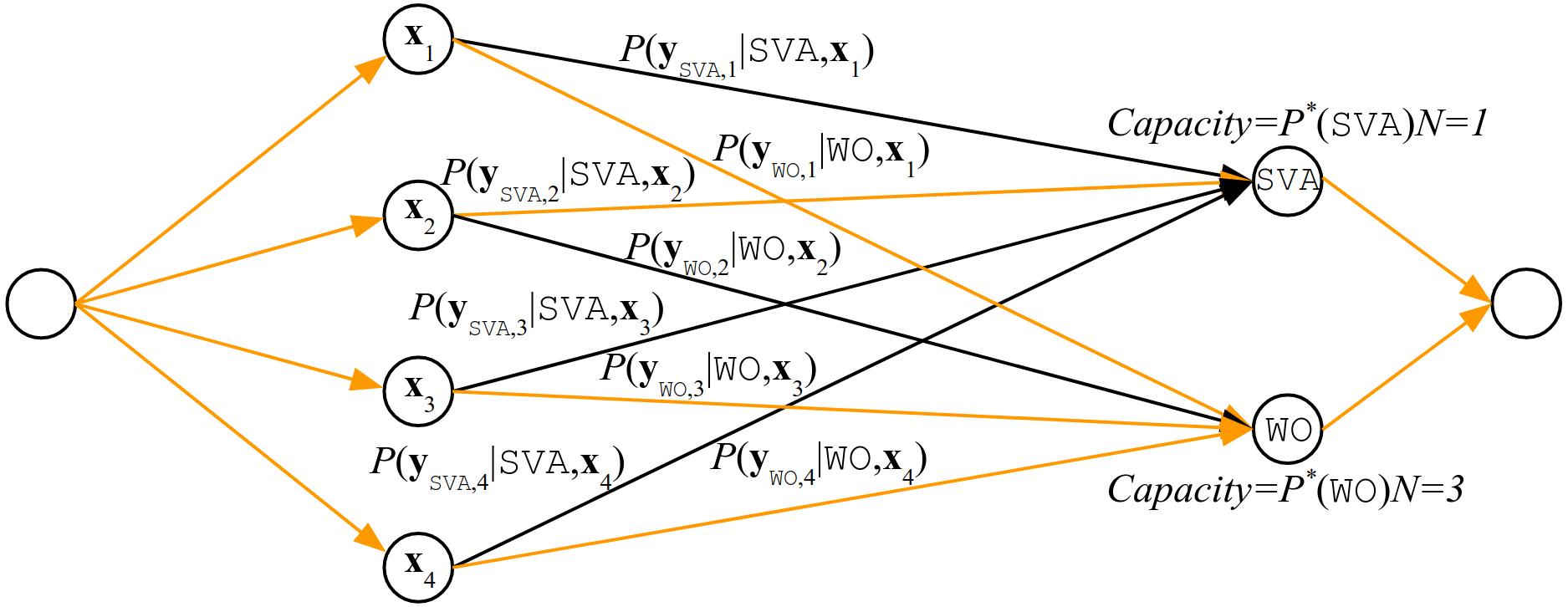}
\caption{The minimum-cost flow graph of the {\em Offline-Optimal} method for a training set of $N=4$ sentences, two error type tags (\texttt{SVA}: subject-verb agreement, and \texttt{WO}: word order) and the desired tag distribution $P^*(\mathtt{SVA})=0.25$ and $P^*(\mathtt{WO})=0.75$. Each sentence is represented by a node on the left side with a flow capacity of one. Each tag is represented by a node on the right side with a capacity equal to the absolute number of sentences for each tag $P^*(t)N$. Arcs connecting sentences and tags are weighted by the corruption model score $P(\mathbf{y}_{t,n}|t,\mathbf{x}_n)$. A possible flow through the graph is highlighted in orange, this assigns the \texttt{WO} tag to the sentences $\mathbf{x}_1$, $\mathbf{x}_3$, $\mathbf{x}_4$ and the \texttt{SVA} tag to the sentence $\mathbf{x}_2$.}
\label{fig:mincostflow}
\end{figure*}

Alternatively, the recently proposed Seq2Edits\footnote{A short description of the Seq2Edits model is provided in Appendix \ref{sec:seq2edits} for convenience.} \citep{seq2edits} model is able to directly predict error tags along with the edits, and does not need to be provided error tags in the input sequence. Instead, during beam search we constrain the tag output tape of Seq2Edits with an FST that forces the generation of a certain tag. Fig.~\ref{fig:constraint-fsts} illustrates three types of constraint FSTs with the example tag, \texttt{SPELL}. All FSTs require at least one occurrence of the \texttt{SPELL} tag. {\sc NoSigma} (Fig.~\ref{fig:no-sigma}) is the most restrictive constraint as it only allows \texttt{SPELL} and \texttt{SELF} (used by Seq2Edits for unmodified source spans). {\sc PostSigma} (Fig.~\ref{fig:post-sigma}) allows other tags after \texttt{SPELL}, but constrains the hypothesis to start with either \texttt{SELF} or \texttt{SPELL} to prevent beam search from committing to a corruption that is incompatible with \texttt{SPELL}.\footnote{An example of this garden-path problem would be a subject-verb-agreement (\texttt{SVA}) constraint, but all active hypotheses in the beam already contain an adjective error and the correct subject and verb (e.g.: ``SVA He owns a large bike with tiny wheels'' $\rightarrow$ ``He owns a wide bike with\dots'').} {\sc PrePostSigma} (Fig.~\ref{fig:pre-post-sigma}) allows other tags both before and after \texttt{SPELL}.

Table \ref{tab:sheep-corruptions} lists example outputs of a tagged Seq2Edits corruption model for all 25 ERRANT tags and demonstrates the model's capability to generate a broad variety of realistic errors.

\section{Synthetic Data Generation with Tagged Corruption Models}

For a grammatical input sentence $\mathbf{x}_n$ ($n\in[1,N]$ where $N$ is the training set size), we denote the corrupted sentence according to the tag $t\in \mathcal{T}$ as $\mathbf{y}_{t,n}$. Our goal is to assign a single tag $t^*_n$ to each training sentence such that the overall distribution follows a certain desired tag distribution $P^*(t)$:
\begin{equation}
\forall t\in\mathcal{T}: P^*(t) \approx \frac{|\{t^*_n = t|n\in[1,N]\}|}{N}
\label{eq:tag-distribution}
\end{equation}
We compare three different methods: {\em Offline-Optimal}, {\em Offline-Probabilistic}, and {\em Online}.

\paragraph{Offline-Optimal}

The {\em Offline-Optimal} method frames this task as a constrained optimization problem:
\begin{equation}
\max_{\mathbf{t}^*} \sum_{n=1}^N \log P(\mathbf{y}_{t^*_n,n}|t^*_n, \mathbf{x}_n)
\end{equation}
under the constraint that the observed distribution of tags matches the desired distribution, i.e.\ Eq.~\ref{eq:tag-distribution} is satisfied. Fig.\ \ref{fig:mincostflow} illustrates that this is an instance of a well-studied problem called maximum weighted bipartite matching~\cite{schrijver-book} and can be solved efficiently with a standard minimum-cost flow solver.

\begin{table*}
\small
\centering
\begin{tabular}{lccc}
\hline
 \textbf{Data set} & \textbf{Synthetic} & \textbf{Number of sentences} & \textbf{Used for}  \\
\hline
WikiEdits \citep{lichtarge-etal-2019-corpora} & No & 170M & Pre-training \\
RoundTripGerman \citep{lichtarge-etal-2019-corpora} & Yes & 176M & Pre-training \\
Lang-8 \citep{mizumoto-etal-2011-mining} & No & 1.9M & Stage 1 fine-tuning \\
FCE-train \citep{yannakoudakis-etal-2011-new} & No & 26K & Stage 2 fine-tuning \\
BEA-train \citep{bryant-etal-2019-bea} & No & 34K & Stage 2 fine-tuning \\
\hline
\textbf{This work:} C4$_\text{200M}$ & Yes & 200M & Pre-training \\
\hline
\end{tabular}
\caption{\label{tab:data} Training data sets used in this work.}
\end{table*}

\paragraph{Offline-Probabilistic}
The intuition behind the {\em Offline-Probabilistic} method is to first draw a tag according to the desired tag distribution $P^{\ast}(t)$ and then sample sentences which are most likely to contain this tag, i.e. draw $N$ sentences from the distribution $P((\mathbf{x},\mathbf{y})|t)$. 
\begin{align*}
    P((\mathbf{x},\mathbf{y})|t) & = & \frac{P(\mathbf{x},\mathbf{y}) P(t|\mathbf{x},\mathbf{y})}{ \sum_{n=1}^{N} P(\mathbf{x}_n,\mathbf{y}_n) P(t|\mathbf{x}_n,\mathbf{y}_n)} \\
\nonumber     & = &  \frac{ P(t|\mathbf{x},\mathbf{y})}{ \sum_{n=1}^{N} P(t|\mathbf{x}_n,\mathbf{y}_n)},
\end{align*}
where we assume each sentence-pair has the same probability $P(\mathbf{x},\mathbf{y}) = \frac{1}{N}$. 
\begin{align*}
    P(t|(\mathbf{x},\mathbf{y})) & = & \frac{P(t,\mathbf{x},\mathbf{y})}{P(\mathbf{x},\mathbf{y})} \\
\nonumber      & = & \frac{P(\mathbf{x}) P(t|\mathbf{x}) P(\mathbf{y}|t,\mathbf{x})}{P(\mathbf{x},\mathbf{y})} \\
\nonumber      & \approx & \frac{\frac{1}{N} \frac{1}{|\mathcal{T}|} P(\mathbf{y}|t,\mathbf{x})}{\frac{1}{N}} \\
\nonumber      & = & \frac{1}{|\mathcal{T}|} P(\mathbf{y}|t,\mathbf{x}), 
\end{align*}
where we assume that a) each sample has equal probability i.e. $P(\mathbf{x}) \approx P(\mathbf{x},\mathbf{y}) = \frac{1}{N}$, b) each tag is equally likely given the source sentence, $P(t|\mathbf{x}) = \frac{1}{|\mathcal{T}|}$, where $|\mathcal{T}|$ is the size of the tag vocabulary. $P(\mathbf{y}|t,\mathbf{x})$ is the probability assigned by the corruption model to the target sequence $\mathbf{y}$ given the source $\mathbf{x}$ and tag $t$.

Unlike the {\em Offline-Optimal} approach, this method does not guarantee that each sentence from the original training set will be included in the sample. However, this may not matter in practice when drawing from a large pool of examples.

\paragraph{Online}

A major limitation that prevents the {\em Offline-Optimal} and {\em Offline-Probabilistic} methods from scaling up efficiently is that we need to run the corruption model for every combination of tag and source sentence ($\Omega(N|\mathcal{T}|)$ runtime).\footnote{We ignore the runtime of beam search when describing the asymptotic time complexity for simplification.} The {\em Online} method avoids this computational complexity by 
drawing the tag $t^*_n$ for each example from the desired tag distribution $P^*(\cdot)$, and then generating the target $\mathbf{y}_n$ given the source $\mathbf{x}_n$ and tag $t^*_n$. 
\begin{equation}
    \forall n\in [1,N]: t^*_n \sim P^*.
\end{equation}
Thus, it does not rely on the corruption model probabilities.  The {\em Online} method assigns tags on-the-fly to each sentence independently and hence runs in $\Theta(N)$.

\section{Results}

For comparability to related work, we report span-based ERRANT $F_{0.5}$-scores on the development and test sets (BEA-dev and BEA-test) of the BEA-2019 shared task \citep{bryant-etal-2019-bea}. We use the M2 scorer \citep{dahlmeier-ng-2012-better} to compute $F_{0.5}$-scores on the CoNLL-13 \citep{ng-etal-2013-conll} and CoNLL-14 \citep{ng-etal-2013-conll} sets, and the GLEU metric \citep{napoles-etal-2015-ground} on JFLEG-dev and JFLEG-test \citep{napoles-etal-2017-jfleg}.

\subsection{Training Setup}
\label{sec:training-setup}

All our grammar {\em correction} models are standard Seq2Seq ({\em not} Seq2Edits) Transformers \citep{transformer} trained with Adafactor \citep{adafactor} using the Tensor2Tensor \citep{vaswani-etal-2018-tensor2tensor} TensorFlow \citep{tensorflow2015-whitepaper} library. Our {\em corruption} models are either standard Transformers or Seq2Edits models \citep{seq2edits}.\footnote{The focus of our work was to examine techniques for synthetic data correction while keeping the {\em correction} model fixed. Hence, we do not use Seq2Edits models for {\em correction}.} We use a Tensor2Tensor joint 32K subword vocabulary and the `Big' hyper-parameter set for all our models. For our tagged corruption models we extend the subword vocabulary by the 25 ERRANT error tags.

We use both existing and new data sets to train our models (Table \ref{tab:data}). WikiEdits and RoundTripGerman are large but noisy pre-training sets described by \citet{lichtarge-etal-2019-corpora}. In this work we introduce a new synthetic pre-training corpus -- C4$_\text{200M}$ -- that we generated by applying our corruption methods to 200M sentences sampled randomly from the Colossal Clean Crawled Corpus \citep[C4]{c4}.\footnote{We filtered C4 with language ID and removed sentences longer than 250 words before selecting the 200M sentences.} Our final {\em correction} models are trained using the two stage fine-tuning recipe of \citet{jared-tacl}: after pre-training we first fine-tune on Lang-8 \citep{mizumoto-etal-2011-mining} and then on BEA+FCE which is the combination of the FCE corpus \citep{yannakoudakis-etal-2011-new} and the training split of the Cambridge English Write \& Improve corpus used in the BEA-2019 shared task \citep{bryant-etal-2019-bea}. Our {\em corruption} models are trained using a similar setup but do not use C4$_\text{200M}$ in pre-training. In our ablation experiments, however, we modify specific stages of this training pipeline to gain more insight into our methods.

\begin{table*}
\small
\centering
\begin{tabular}{@{\hspace{0em}}l@{\hspace{0.2em}}lccccccccccc}
\hline
& \textbf{Corruption model} & \textbf{Constraint} & \textbf{Tagged} & \multicolumn{3}{c}{\textbf{Offline-Optimal}} & \multicolumn{3}{c}{\textbf{Offline-Probabilistic}} & \multicolumn{3}{c}{\textbf{Online}} \\
& & & \textbf{input?} & \textbf{P} & \textbf{R} & \textbf{F$_{0.5}$} & \textbf{P} & \textbf{R} & \textbf{F$_{0.5}$} & \textbf{P} & \textbf{R} & \textbf{F$_{0.5}$} \\
\hline
\footnotesize{a} & Full sequence & - & $\checkmark$ & 54.5 & 16.7 & 37.5 & 51.0 & 20.7 & 39.4 & 53.5 & 18.5 & 38.8 \\
\footnotesize{b} & Seq2Edits & \sc{NoSigma} &  & 57.2 & 21.8 & 43.2 & 57.1 & 25.6 & 45.8 & 57.5 & 26.4 & 46.6 \\
\footnotesize{c} & Seq2Edits & \sc{PostSigma} &  & 56.5 & 21.6 & 42.7 & 56.6 & 25.7 & 45.6 & 59.4 & 27.2 & 48.0 \\
\footnotesize{d} & Seq2Edits & \sc{PrePostSigma} &  & 56.3 & 22.4 & 43.3 & 55.7 & 25.5 & 45.0 & 53.0 & 30.6 & 46.2 \\
\footnotesize{e} & Seq2Edits & -  & $\checkmark$ & 55.3 & 26.7 & 45.5 & 55.6 & 25.9 & 45.2 & 54.4 & 29.0 & 46.3 \\
\hline
\end{tabular}
\caption{\label{tab:synth-with-tags} {\em Tagged} synthetic data generation where tags are chosen according to the BEA-dev tag distribution. The GEC seed model (Table \ref{tab:synth-without-tags}a) is fine-tuned on BEA+FCE with synthetic source sentences and evaluated on the BEA-dev set. Rows \textbf{a} and \textbf{e} prepend the desired tag to the input sequence while rows \textbf{b}-\textbf{d} use FST constraints.}
\end{table*}

\begin{table}
\small
\centering
\begin{tabular}{@{\hspace{0em}}l@{\hspace{0.2em}}l@{\hspace{0.6em}}lccc}
\hline
& \multicolumn{2}{c}{\textbf{GEC data (2.\ fine-tuning)}} & \multicolumn{3}{c}{\textbf{BEA-dev}} \\
& \multicolumn{1}{c}{\textbf{Source}} & \multicolumn{1}{c}{\textbf{Target}} & \textbf{P} & \textbf{R} & \textbf{F$_{0.5}$}  \\
\hline
\footnotesize{a} & \multicolumn{2}{c}{N/A (Seed model)} & 57.0 & 12.8 & 33.7 \\
\footnotesize{b} &  Real data & BEA+FCE & 56.5 & 35.2 & 50.4 \\
\footnotesize{c} & Synthetic (Full seq.) & BEA+FCE & 57.6 & 20.6 & 42.4 \\
\footnotesize{d} & Synthetic (Seq2Edits) & BEA+FCE & 53.4 & 20.6 & 40.5 \\
\hline
\end{tabular}
\caption{\label{tab:synth-without-tags} {\em Untagged} synthetic data generation. The seed GEC model (row \textbf{a}) is fine-tuned on BEA+FCE target sentences that are either paired with the real source sentences (row \textbf{b}) or with back-translated source sentences using either a full sequence Transformer (row \textbf{c}) or a Seq2Edits (row \textbf{d}) corruption model.}
\end{table}

\subsection{Synthetic vs.\ Real Parallel Data}
\label{sec:synth-vs-real}

In initial experiments (Tables \ref{tab:synth-with-tags} to \ref{tab:cefr}) we explore how well our synthetic data generation methods can replace real parallel data. The corruption models used in this section are fine-tuned on Lang-8 but not on BEA+FCE. The seed correction model is pre-trained on WikiEdits and RoundTripGerman and fine-tuned on Lang-8. We discard the source sentences in BEA+FCE, replace them with synthetic corruptions of the target sentences, and fine-tune the seed correction model on this synthetic data, i.e.\ all models in Tables \ref{tab:synth-with-tags} to \ref{tab:cefr} are trained by fine-tuning the same seed model (Table \ref{tab:synth-without-tags}a and \ref{tab:cefr}a) on the same set of target sentences but different sets of source sentences.

\paragraph{Data generation without tags}

Fine-tuning the seed model on the real parallel data improves the $F_{0.5}$-score on BEA-dev by 16.7 points (33.7 $\rightarrow$ 50.4 in rows \textbf{a} and \textbf{b} of Table \ref{tab:synth-without-tags}). Our goal is to close the gap relative to the $F_{0.5}$ of 50.4 using synthetic source sentences. The corruption models in rows \textbf{c} and \textbf{d} of Table \ref{tab:synth-without-tags} do not use any error tags, which is similar to previous attempts to apply back-translation to GEC \citep{kasewa-etal-2018-wronging}. 

\paragraph{Data generation with tags}

Table \ref{tab:synth-with-tags} reports results from the tag-based corruption methods introduced in this work. Seq2Edits (rows \textbf{b}-\textbf{e}) is more amenable to tag-based corruption than a standard full sequence Transformer model (row \textbf{a}) because tag prediction is a component of the Seq2Edits model. Interestingly, the {\em Offline-Optimal} method tends to perform worse than {\em Offline-Probabilistic} and {\em Online} in the constrained Seq2Edits experiments (rows \textbf{b}-\textbf{e}). We hypothesize that {\em Offline-Optimal} might generate duller and more systematic errors because the corruption model score is used to pair tags with sentences. Increasing the diversity of synthetic errors by selecting non-optimal tag-sentence pairs ultimately improves the usefulness of the synthetic data.\footnote{The same intuition motivates our experiments in Sec.~\ref{sec:c4200m} that replace beam search with sampling.}

Comparing Table \ref{tab:synth-without-tags} with Table \ref{tab:synth-with-tags} we observe that controlling the tag distribution of the synthetic data from a Seq2Edits model outperforms traditional back-translation without tags. Our best model ({\em Online} column in Table \ref{tab:synth-with-tags}c) achieves an $F_{0.5}$-score of 48.0 which is much better than our best system without tags (42.4 $F_{0.5}$ in Table \ref{tab:synth-without-tags}c) and remarkably close to the oracle score of 50.4 $F_{0.5}$ (Table \ref{tab:synth-without-tags}b) obtained using a model trained on real parallel data.

\begin{table}[t!]
\small
\centering
\begin{tabular}{@{\hspace{0em}}l@{\hspace{0.2em}}lcccc}
\hline
& \textbf{System} & \multicolumn{4}{c}{\textbf{Test set (F$_{0.5}$)}} \\
&  & \textbf{A2} & \textbf{B2} & \textbf{C2} & \textbf{N2}\\
\hline
 \multicolumn{6}{l}{\textbf{Baselines}} \\
\hline
\footnotesize{a} & Seed model & 37.6 & 34.1 & 31.4 & 22.2 \\
\footnotesize{b} & FT on real data  & \textbf{50.3} & \textbf{51.5} & \textbf{44.1} & 42.1 \\
\hline
 \multicolumn{6}{l}{\textbf{Synthetic data using target tag distributions $P^*(t)$}} \\
\hline
\footnotesize{c} & CEFR-A (A1) & 47.4 & 46.2 & 39.0 & 39.0   \\
\footnotesize{d} & CEFR-B (B1) & 47.1 & 46.0 & 40.9 & 38.0   \\
\footnotesize{e} & CEFR-C (C1) & 47.1 & 46.2 & 37.1 & 39.1   \\
\footnotesize{f} & Native (N1) & 47.8 & 49.2 & 42.8 & \textbf{42.9}   \\
\hline
\end{tabular}
\caption{\label{tab:cefr} Adapting GEC to non-native or native English. In rows \textbf{c}-\textbf{f} the GEC seed model (row \textbf{a}) is fine-tuned on BEA+FCE with source sentences synthesized by a tagged Seq2Edits corruption model by following proficiency-dependent tag distributions (A1, B1, C1, or N1). FT denotes fine-tuning.}
\end{table}

\begin{figure*}[t!]
\centering
\small
\includegraphics[scale=0.9]{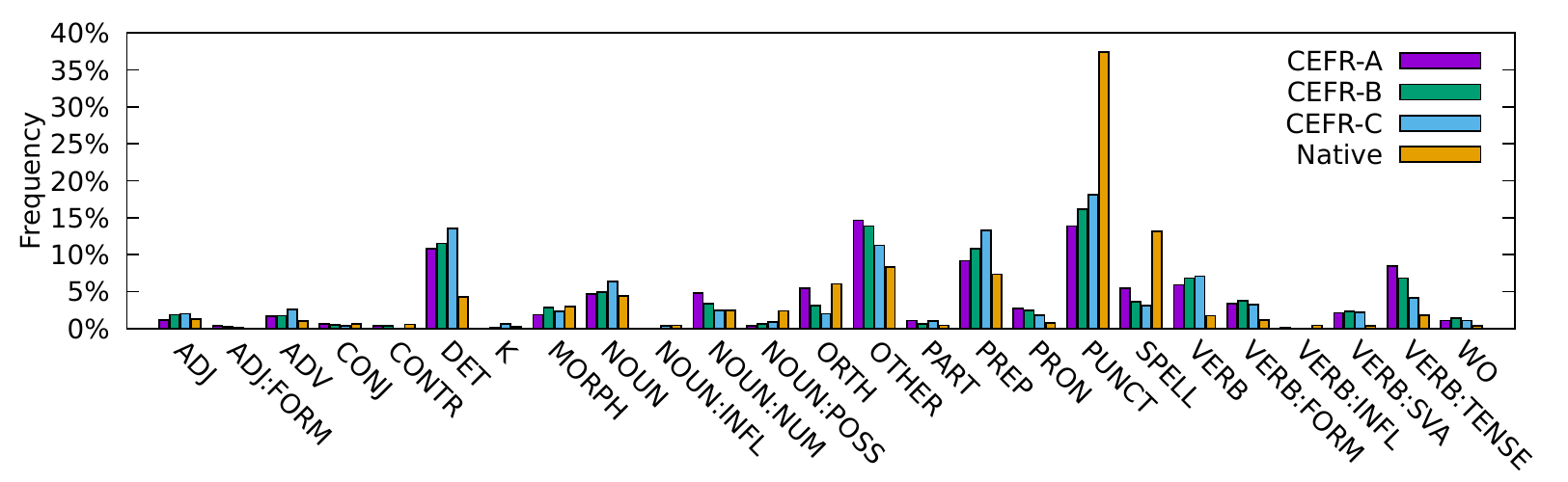}
\caption{ERRANT error tag distributions on BEA-dev for non-native (CEFR levels A, B, C) and native English.}
\label{fig:distributions}
\end{figure*}

\begin{figure*}[t!]
\centering
\small
\includegraphics[scale=0.9]{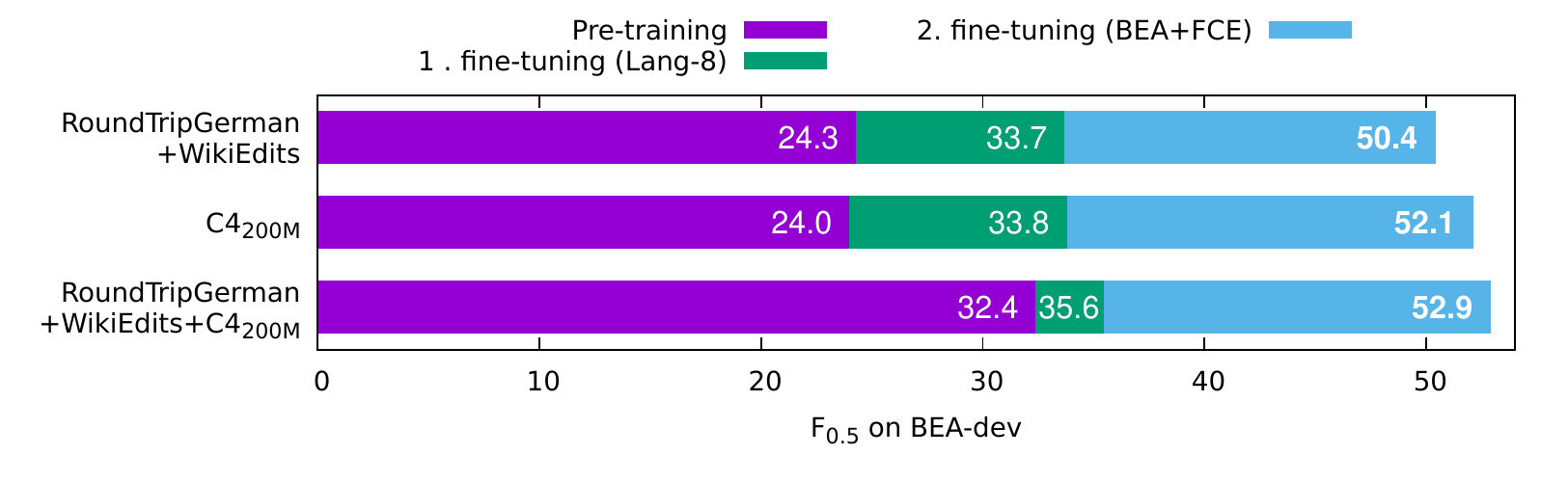}
\caption{Using C4$_\text{200M}$ (tagged Seq2Edits corruption model, BEA-dev tag distribution) in pre-training. The 2-stage fine-tuning setup is described in Sec.~\ref{sec:training-setup}.}
\label{fig:pre-bar-chart}
\end{figure*}

For all experiments in the remainder of this paper we used the unconstrained tagged Seq2Edits corruption models (Table \ref{tab:synth-with-tags}e)  because it yields reasonable gains across all methods ({\em Offline-Optimal}, {\em Offline-Probabilistic}, and {\em Online}) and is easiest and most practical to run on a large scale.\footnote{We noticed that constrained decoding (Table \ref{tab:synth-with-tags}b-d) often fails in large-scale experiments if the selected tag and source sentence are incompatible.} Furthermore, we will only use the {\em Online} method to avoid the computational overhead of {\em Offline-Optimal} and {\em Offline-Probabilistic}.

\paragraph{Adapting GEC to English proficiency levels}

A potential use case for tagged corruption models is to adapt a GEC model to the proficiency level of the user by changing the target tag distribution $P^*(t)$ of the synthetic fine-tuning set. Each sentence in the BEA-dev development set is annotated with English proficiency labels (CEFR-levels A, B, C, and `N' for native English). We split BEA-dev using these labels, and then split each set again into two parts (development/test), resulting in eight disjoint subsets A1, A2, B1, B2, C1, C2, N1, N2 with about 500 sentences each. We use A1, B1, C1, and N1 to estimate proficiency-dependent target tag distributions. As before we fine-tune the seed model on the BEA-train target sentences with synthetic source sentences that follow one of these tag distributions, but evaluate the fine-tuned models on the A2, B2, C2, and N2 splits. Table \ref{tab:cefr} shows that the tag distributions from A1, B1, and C1 yield similar performance (rows \textbf{c}-\textbf{e} in Table \ref{tab:cefr}) across most test sets. This suggests that our method is not effective at discriminating between the different CEFR-levels of non-native English. However, using the tag distribution from native speakers (N1 in Table \ref{tab:cefr}f) does yield substantial gains on the native English test set (42.9 $F_{0.5}$ on N2), even surpassing the real parallel data (42.1 $F_{0.5}$ Table \ref{tab:cefr}b). This demonstrates the potential of tag-based corruption for improving GEC of native English.

Fig.~\ref{fig:distributions} shows that the error tag distribution for native English differs significantly from the non-native distributions. Native speakers (orange bar) tend to make more punctuation (\texttt{PUNCT}), and spellling (\texttt{SPELL}) mistakes whereas the determiner errors (\texttt{DET}) are more common in non-native text.

\subsection{The C4$_\text{200M}$ Synthetic Data Set}
\label{sec:c4200m}

\begin{table*}
\small
\centering
\begin{tabular}{@{\hspace{0em}}l@{\hspace{0.2em}}llcccccccccc}
\hline
& \textbf{Tag distribution}  & \textbf{Decoding} & \multicolumn{3}{c}{\textbf{BEA-dev}} & \multicolumn{3}{c}{\textbf{CoNLL-13}} & \textbf{JFLEG-dev} \\
& \textbf{$P^*(t)$} & & \textbf{P} & \textbf{R} & \textbf{F$_{0.5}$} & \textbf{P} & \textbf{R} & \textbf{F$_{0.5}$} & \textbf{GLEU} \\
\hline
\footnotesize{a} & None (no tags) & Beam search & 58.1 & 36.5 & 51.9 & 58.5 & 29.0 & 48.6 & 57.2 \\
\footnotesize{b} & BEA-dev & Beam search & 58.0 & 39.3 & 52.9 & 58.5 & 33.0 & 50.7 & 57.9 \\
\footnotesize{c} & CoNLL-13 & Beam search & 58.8 & 39.6 & 53.6 & 58.9 & 33.3 & 51.0 & 57.9 \\
\footnotesize{d} & JFLEG-dev & Beam search & 58.3 & 39.1 & 53.1 & 58.9 & 31.6 & 50.2 & 57.6 \\
\footnotesize{e} & Uniform & Beam search & 59.1 & 39.6 & 53.8 & 58.3 & 33.4 & 50.7 & 57.7 \\
\hline
\footnotesize{f} & None (no tags) & Sampling & 57.5 & 36.0 & 51.4 & 56.9 & 29.4 & 47.9 & 57.1 \\
\footnotesize{g} & BEA-dev & Sampling & 59.5 & \textbf{41.3} & \textbf{54.7} & \textbf{59.2} & \textbf{34.7} & \textbf{51.9} & \textbf{58.5} \\
\footnotesize{h} & CoNLL-13 & Sampling & 58.6 & 40.7 & 53.9 & 58.5 & 33.3 & 50.8 & 58.1 \\
\footnotesize{i} & JFLEG-dev & Sampling & 59.0 & 39.7 & 53.8 & 58.2 & 34.0 & 51.0 & 58.4 \\
\footnotesize{j} & Uniform & Sampling & \textbf{59.6} & 40.6 & 54.5 & 58.3 & 34.3 & 51.1 & 58.3 \\
\hline
\end{tabular}
\caption{\label{tab:cross-dev-set} Using different target tag distributions to corrupt C4$_\text{200M}$ with a tagged Seq2Edits corruption model, either using beam search or sampling. We report the best of five training runs after two-stage fine-tuning according to the performance on the development set.}
\end{table*}

\begin{table}[t!]
\small
\centering
\begin{tabular}{@{\hspace{0em}}l@{\hspace{0.2em}}l@{\hspace{0.6em}}lccc}
\hline
& \multicolumn{2}{c}{\textbf{GEC data (pre-training)}} & \multicolumn{3}{c}{\textbf{BEA-dev}} \\
& \multicolumn{1}{c}{\textbf{Synthetic source}} & \multicolumn{1}{c}{\textbf{Target}} & \textbf{P} & \textbf{R} & \textbf{F$_{0.5}$}  \\
\hline
\footnotesize{a} & Untagged corruption & WikiEdits & 40.8 & 4.0 & 14.3 \\
\footnotesize{b} & RoundTripGerman & WikiEdits & 33.6 & 11.6 & 24.3 \\
\footnotesize{c} & Tagged corruption & WikiEdits & 39.7 & 20.2 & 33.3 \\
\hline
\end{tabular}
\caption{\label{tab:wikicorrupt} Data generation on WikiEdits for pre-training. Models are pre-trained on a mix of the original WikiEdits data set and a synthetic data set that consists of WikiEdits target sentences corrupted with either (1) an untagged Seq2Edits corruption model (row \textbf{a}), (2) round-trip translation via German (row \textbf{b}, or row 1 in Fig.~\ref{fig:pre-bar-chart}), or (3) a tagged Seq2Edits corruption model (row \textbf{c}) following the BEA-dev tag distribution.}
\end{table}

We showed in the previous section that using an unconstrained tagged Seq2Edits corruption model that follows the BEA-dev tag distribution works well in a controlled setup (corrupting $\sim$60K clean target sentences from BEA+FCE). We now apply the same corruption model to a much larger, clean  data set (C4$_\text{200M}$) consisting of 200M sentences and use the resulting synthetic data set as an additional pre-training set for our GEC models. Fig.~\ref{fig:pre-bar-chart} reports performance from three different GEC models with different pre-training sets, each using the 2-stage fine-tuning pipeline described in Sec.~\ref{sec:training-setup}. The RoundTripGerman+WikiEdits model resembles the baseline of \citet{jared-tacl}. Using C4$_\text{200M}$ instead of RoundTripGerman+WikiEdits improves the final F$_{0.5}$-score to 52.1. Combining all three pre-training sets leads to a large jump in F$_{0.5}$ to 32.4 after pre-training. The gains are reduced after fine-tuning, but our best model still uses all three pre-training sets (52.9 F$_{0.5}$ after the second fine-tuning stage).
The gains in Fig.~\ref{fig:pre-bar-chart} from using C4$_\text{200M}$ can be attributed to a) the tagged corruption method, or b) the use of C4 rather than Wikipedia which covers a broader range of text types. In the ablation experiment in Table \ref{tab:wikicorrupt}, rather than using C4$_\text{200M}$, we corrupted the WikiRevision target sentences with various corruption methods. Tagged corruption (row \textbf{c}) outperforms both untagged corruption (row \textbf{a}) and round-trip translation (row \textbf{b}) when the target sentences are kept constant.

\begin{table*}
\small
\centering
\begin{tabular}{lcccccccccc}
\hline
\textbf{System}  & \multicolumn{3}{c}{\textbf{BEA-test}} & \multicolumn{3}{c}{\textbf{CoNLL-14}} & \textbf{JFLEG-test} \\
 & \textbf{P} & \textbf{R} & \textbf{F$_{0.5}$} & \textbf{P} & \textbf{R} & \textbf{F$_{0.5}$} & \textbf{GLEU} \\
\hline
\textbf{Single systems} &  &  &  &  &  &  &   \\
\hline
\citet{gec-pseudo-data} & 65.5 & 59.4 & 64.2 & 67.9 & 44.1 & 61.3 & 59.7 \\
\citet{gector} & \textbf{79.2} & 53.9 & \textbf{72.4} & \textbf{77.5} & 40.1 & 65.3 & -  \\
\citet{jared-tacl} & 67.6 & 62.5 & 66.5 & 69.4 & 43.9 & 62.1 & 63.8 \\
\citet{kaneko-etal-2020-encoder} & 67.1 & 60.1 &  65.6 & 69.2 & 45.6 & 62.6 & 61.3 \\
\citet{wan-etal-2020-improving} & 66.9 & 60.6 & 65.5 & 69.5 & 47.3 & 63.5 & - \\
\textbf{This work} & 72.1 & \textbf{64.4} & 70.4 & 72.8 & \textbf{49.5} & \textbf{66.6} & \textbf{64.7} \\
\hline
\textbf{Ensembles} &  &  &  &  &  &  &   \\
\hline
\citet{grundkiewicz-etal-2019-neural} & 72.3 & 60.1 & 69.5 & - & - & 64.2 & 61.2 \\
\citet{gec-pseudo-data} & 74.7 & 56.7 & 70.2 & 72.4 & 46.1 & 65.0 & 61.4 \\
\citet{gector} & \textbf{79.4} & 57.2 & 73.7 & \textbf{78.2} & 41.5 & 66.5 & - \\
\citet{jared-tacl} & 75.4 & 64.7 & 73.0 & 74.7 & 46.9 & 66.8 & \textbf{64.9} \\
\citet{kaneko-etal-2020-encoder} & 72.3 & 61.4 & 69.8 & 72.6 & 46.4 & 65.2 & 62.0 \\
\citet{wan-etal-2020-improving} & 72.6 & 61.3 & 70.0 & 72.3 & 48.8 & 65.9 & - \\
\textbf{This work} & 77.7 & \textbf{65.4} & \textbf{74.9} & 75.6 & \textbf{49.3} & \textbf{68.3} & 64.7 \\
\hline
\end{tabular}
\caption{\label{tab:comparison-literature} Comparison of our final system with related work.}
\end{table*}

A crucial practical question is whether our approach is sensitive to the particular target tag distribution $P^*(t)$, and if the synthetic C4$_\text{200M}$ training data can help generalization to other development sets. Rows \textbf{b}-\textbf{e} in Table \ref{tab:cross-dev-set} show the performance after fine-tuning for four different tag distributions: BEA-dev, CoNLL-13, JFLEG-dev, and Uniform. Each row reports the performance of a model pre-trained using RoundTripGerman+WikiEdits and C4$_\text{200M}$ corrupted using the desired tag distribution followed by the 2-stage fine-tuning, i.e.\ row \textbf{b} corresponds to row 3 in Fig.\ \ref{fig:pre-bar-chart}. 
All tagged corruption models improve upon the untagged models (rows \textbf{a} and \textbf{f}).\footnote{For more insight into the difference between untagged and tagged corruptions see Appendix \ref{sec:example-corruptions}.} In contrast to our adaptation experiments in Table \ref{tab:cefr}, the variations between different tag distributions are small. This indicates that even though choosing the {\em correct} tag distribution is crucial for adapting GEC to native English, at the pre-training stage the ability of tagged corruption models to generate diverse errors is more important than matching a particular distribution.

Previous work on back-translation has found that it can be beneficial to use sampling instead of beam search for synthetic data generation \citep{edunov-etal-2018-understanding,gec-pseudo-data}. We confirm these findings for our tagged corruption models: Sampling (Table \ref{tab:cross-dev-set}g-j) outperforms beam search (Table \ref{tab:cross-dev-set}b-e) for all tag distributions except CoNLL-13.
Using sampling and the BEA-dev tag distribution (Table \ref{tab:cross-dev-set}g) yields good performance across all development sets. The BEA-dev tag distribution reflects a wide range of grammatical errors across various proficiency levels compared to other corpora such as CoNLL-14 (mostly beginner) or FCE (School) \citep{bryant-etal-2019-bea}. Table \ref{tab:comparison-literature} situates this single model and an ensemble of five analogously trained models in the context of related work. For our final models in Table \ref{tab:comparison-literature} we follow \citet{lichtarge-etal-2019-corpora,jared-tacl,seq2edits} and multiply the model score of the identity mapping with a factor (tuned on the development set) to balance precision and recall.\footnote{This factor is around 1.0 for BEA-dev and CoNLL-13 (i.e.\ no impact) but it helps to re-balance precision and recall on JFLEG (around 2.0).} Our single model outperforms other single models on CoNLL-14 and JFLEG-test. Our ensemble establishes new state-of-the-art scores on BEA-test (74.9 F$_{0.5}$) and CoNLL-14 (68.3 F$_{0.5}$). We would like to emphasize that these gains are achieved without modifying the GEC model architecture -- our GEC models are vanilla Transformers that were pre-trained using our new synthetic C4$_\text{200M}$ data set. We will make our data set publicly available to make it easy for other researchers to benefit from our work. Appendix \ref{sec:example-outputs} contains example outputs from our system trained with C4$_\text{200M}$ that demonstrate improved fluency and better handling of long-range reorderings.

\section{Related Work}

The body of literature on synthetic data generation for GEC is large. Various heuristics have been proposed to inject synthetic noise into grammatical sentences such as random word- or character-level insertion, substitution, deletion, or shuffling operations \citep{lichtarge-etal-2019-corpora,zhao-etal-2019-improving,xu-etal-2019-erroneous}, using spell checkers \citep{grundkiewicz-etal-2019-neural}, or randomly applying word edits extracted from the training data \citep{choe-etal-2019-neural}. \citet{kasewa-etal-2018-wronging,stahlberg-byrne-2019-cueds} applied back-translation \citep{sennrich-etal-2016-improving} to GEC and reported substantial gains. Similar to MT~\citep{edunov-etal-2018-understanding}, back-translation for GEC can be further improved by adding noise to the decoding process \citep{xie-etal-2018-noising} or by using sampling instead of beam search \citep{gec-pseudo-data}. Fluency-boost learning \citep{ge-etal-2018-fluency,ge-fluency} can also be used to generate additional sentence pairs during training. \citet{lichtarge-etal-2019-corpora} proposed to generate noisy counterparts of grammatical English sentences by translating them to another language (e.g.\ German) and back (``round-trip translation''), a technique we also use in this work. The use of tags for back-translation in MT has been explored by \citet{caswell-etal-2019-tagged}. Our tagged corruption models are inspired by \citet{wan-etal-2020-improving} who generated synthetic sentences from latent representations that are perturbed using explicit error type tags. Our approach of adding the tags to the input sequence is simpler as it requires no modifications to the model architecture or training procedure.

\section{Conclusion}

We have introduced a synthetic data generation method for grammatical error correction that is able to produce a wide range of realistic grammatical errors. Our method is based on grammar corruption models that corrupt a clean sentence given an error type tag. Conditioning on the error type tag enables us to control synthetic data generation much more precisely than alternative methods such as round-trip translations or tag-independent back-translation. 
We explored different ways of using these tagged corruption models to generate synthetic data that follows a certain error tag distribution. We found that fine-tuning a model on synthetic data that follows a native English error tag distribution can even outperform fine-tuning on genuine parallel data from a mixture of proficiency levels.
Along with this paper we \href{https://github.com/google-research-datasets/C4_200M-synthetic-dataset-for-grammatical-error-correction}{published a new 200M sentence data set for GEC -- C4$_\text{200M}$}. Using C4$_\text{200M}$ in pre-training of vanilla Transformer GEC models yields state-of-the-art performance on two standard GEC test sets (BEA-test and CoNLL-14). We expect this corpus to further stimulate the development of new data-driven approaches in GEC.

\bibliography{anthology,eacl2021}

\begin{thebibliography}{43}
\expandafter\ifx\csname natexlab\endcsname\relax\def\natexlab#1{#1}\fi

\bibitem[{Abadi et~al.(2015)Abadi, Agarwal, Barham, Brevdo, Chen, Citro,
  Corrado, Davis, Dean, Devin, Ghemawat, Goodfellow, Harp, Irving, Isard, Jia,
  Jozefowicz, Kaiser, Kudlur, Levenberg, Man\'{e}, Monga, Moore, Murray, Olah,
  Schuster, Shlens, Steiner, Sutskever, Talwar, Tucker, Vanhoucke, Vasudevan,
  Vi\'{e}gas, Vinyals, Warden, Wattenberg, Wicke, Yu, and
  Zheng}]{tensorflow2015-whitepaper}
Mart\'{\i}n Abadi, Ashish Agarwal, Paul Barham, Eugene Brevdo, Zhifeng Chen,
  Craig Citro, Greg~S. Corrado, Andy Davis, Jeffrey Dean, Matthieu Devin,
  Sanjay Ghemawat, Ian Goodfellow, Andrew Harp, Geoffrey Irving, Michael Isard,
  Yangqing Jia, Rafal Jozefowicz, Lukasz Kaiser, Manjunath Kudlur, Josh
  Levenberg, Dandelion Man\'{e}, Rajat Monga, Sherry Moore, Derek Murray, Chris
  Olah, Mike Schuster, Jonathon Shlens, Benoit Steiner, Ilya Sutskever, Kunal
  Talwar, Paul Tucker, Vincent Vanhoucke, Vijay Vasudevan, Fernanda Vi\'{e}gas,
  Oriol Vinyals, Pete Warden, Martin Wattenberg, Martin Wicke, Yuan Yu, and
  Xiaoqiang Zheng. 2015.
\newblock \href {https://www.tensorflow.org/} {{TensorFlow}: Large-scale
  machine learning on heterogeneous systems}.
\newblock Software available from tensorflow.org.

\bibitem[{Brockett et~al.(2006)Brockett, Dolan, and
  Gamon}]{brockett-etal-2006-correcting}
Chris Brockett, William~B. Dolan, and Michael Gamon. 2006.
\newblock \href {https://doi.org/10.3115/1220175.1220207} {Correcting {ESL}
  errors using phrasal {SMT} techniques}.
\newblock In \emph{Proceedings of the 21st International Conference on
  Computational Linguistics and 44th Annual Meeting of the Association for
  Computational Linguistics}, pages 249--256, Sydney, Australia. Association
  for Computational Linguistics.

\bibitem[{Bryant et~al.(2019)Bryant, Felice, Andersen, and
  Briscoe}]{bryant-etal-2019-bea}
Christopher Bryant, Mariano Felice, {\O}istein~E. Andersen, and Ted Briscoe.
  2019.
\newblock \href {https://doi.org/10.18653/v1/W19-4406} {The {BEA}-2019 shared
  task on grammatical error correction}.
\newblock In \emph{Proceedings of the Fourteenth Workshop on Innovative Use of
  NLP for Building Educational Applications}, pages 52--75, Florence, Italy.
  Association for Computational Linguistics.

\bibitem[{Bryant et~al.(2017)Bryant, Felice, and
  Briscoe}]{bryant-etal-2017-automatic}
Christopher Bryant, Mariano Felice, and Ted Briscoe. 2017.
\newblock \href {https://doi.org/10.18653/v1/P17-1074} {Automatic annotation
  and evaluation of error types for grammatical error correction}.
\newblock In \emph{Proceedings of the 55th Annual Meeting of the Association
  for Computational Linguistics (Volume 1: Long Papers)}, pages 793--805,
  Vancouver, Canada. Association for Computational Linguistics.

\bibitem[{Caswell et~al.(2019)Caswell, Chelba, and
  Grangier}]{caswell-etal-2019-tagged}
Isaac Caswell, Ciprian Chelba, and David Grangier. 2019.
\newblock \href {https://doi.org/10.18653/v1/W19-5206} {Tagged
  back-translation}.
\newblock In \emph{Proceedings of the Fourth Conference on Machine Translation
  (Volume 1: Research Papers)}, pages 53--63, Florence, Italy. Association for
  Computational Linguistics.

\bibitem[{Choe et~al.(2019)Choe, Ham, Park, and Yoon}]{choe-etal-2019-neural}
Yo~Joong Choe, Jiyeon Ham, Kyubyong Park, and Yeoil Yoon. 2019.
\newblock \href {https://doi.org/10.18653/v1/W19-4423} {A neural grammatical
  error correction system built on better pre-training and sequential transfer
  learning}.
\newblock In \emph{Proceedings of the Fourteenth Workshop on Innovative Use of
  NLP for Building Educational Applications}, pages 213--227, Florence, Italy.
  Association for Computational Linguistics.

\bibitem[{Chollampatt and Ng(2018)}]{chollampatt2018multilayer}
Shamil Chollampatt and Hwee~Tou Ng. 2018.
\newblock A multilayer convolutional encoder-decoder neural network for
  grammatical error correction.
\newblock In \emph{Proceedings of the Thirty-Second AAAI Conference on
  Artificial Intelligence}.

\bibitem[{Dahlmeier and Ng(2012)}]{dahlmeier-ng-2012-better}
Daniel Dahlmeier and Hwee~Tou Ng. 2012.
\newblock \href {https://www.aclweb.org/anthology/N12-1067} {Better evaluation
  for grammatical error correction}.
\newblock In \emph{Proceedings of the 2012 Conference of the North {A}merican
  Chapter of the Association for Computational Linguistics: Human Language
  Technologies}, pages 568--572, Montr{\'e}al, Canada. Association for
  Computational Linguistics.

\bibitem[{Edunov et~al.(2018)Edunov, Ott, Auli, and
  Grangier}]{edunov-etal-2018-understanding}
Sergey Edunov, Myle Ott, Michael Auli, and David Grangier. 2018.
\newblock \href {https://doi.org/10.18653/v1/D18-1045} {Understanding
  back-translation at scale}.
\newblock In \emph{Proceedings of the 2018 Conference on Empirical Methods in
  Natural Language Processing}, pages 489--500, Brussels, Belgium. Association
  for Computational Linguistics.

\bibitem[{Felice et~al.(2016)Felice, Bryant, and
  Briscoe}]{felice-etal-2016-automatic}
Mariano Felice, Christopher Bryant, and Ted Briscoe. 2016.
\newblock \href {https://www.aclweb.org/anthology/C16-1079} {Automatic
  extraction of learner errors in {ESL} sentences using linguistically enhanced
  alignments}.
\newblock In \emph{Proceedings of {COLING} 2016, the 26th International
  Conference on Computational Linguistics: Technical Papers}, pages 825--835,
  Osaka, Japan. The COLING 2016 Organizing Committee.

\bibitem[{Felice et~al.(2014)Felice, Yuan, Andersen, Yannakoudakis, and
  Kochmar}]{felice-etal-2014-grammatical}
Mariano Felice, Zheng Yuan, {\O}istein~E. Andersen, Helen Yannakoudakis, and
  Ekaterina Kochmar. 2014.
\newblock \href {https://doi.org/10.3115/v1/W14-1702} {Grammatical error
  correction using hybrid systems and type filtering}.
\newblock In \emph{Proceedings of the Eighteenth Conference on Computational
  Natural Language Learning: Shared Task}, pages 15--24, Baltimore, Maryland.
  Association for Computational Linguistics.

\bibitem[{Foster and Andersen(2009)}]{foster-etal-2009-generate}
Jennifer Foster and {\O}istein~E. Andersen. 2009.
\newblock Generrate: Generating errors for use in grammatical error detection.
\newblock In \emph{Proceedings of the Fourth Workshop on Innovative Use of NLP
  for Building Educational Applications}, EdAppsNLP '09, pages 82--90,
  Stroudsburg, PA, USA. Association for Computational Linguistics.

\bibitem[{Ge et~al.(2018{\natexlab{a}})Ge, Wei, and
  Zhou}]{ge-etal-2018-fluency}
Tao Ge, Furu Wei, and Ming Zhou. 2018{\natexlab{a}}.
\newblock \href {https://doi.org/10.18653/v1/P18-1097} {Fluency boost learning
  and inference for neural grammatical error correction}.
\newblock In \emph{Proceedings of the 56th Annual Meeting of the Association
  for Computational Linguistics (Volume 1: Long Papers)}, pages 1055--1065,
  Melbourne, Australia. Association for Computational Linguistics.

\bibitem[{Ge et~al.(2018{\natexlab{b}})Ge, Wei, and Zhou}]{ge-fluency}
Tao Ge, Furu Wei, and Ming Zhou. 2018{\natexlab{b}}.
\newblock Reaching human-level performance in automatic grammatical error
  correction: An empirical study.
\newblock \emph{arXiv preprint arXiv:1807.01270}.

\bibitem[{Grundkiewicz et~al.(2019)Grundkiewicz, Junczys-Dowmunt, and
  Heafield}]{grundkiewicz-etal-2019-neural}
Roman Grundkiewicz, Marcin Junczys-Dowmunt, and Kenneth Heafield. 2019.
\newblock \href {https://doi.org/10.18653/v1/W19-4427} {Neural grammatical
  error correction systems with unsupervised pre-training on synthetic data}.
\newblock In \emph{Proceedings of the Fourteenth Workshop on Innovative Use of
  NLP for Building Educational Applications}, pages 252--263, Florence, Italy.
  Association for Computational Linguistics.

\bibitem[{Johnson et~al.(2017)Johnson, Schuster, Le, Krikun, Wu, Chen, Thorat,
  Vi{\'e}gas, Wattenberg, Corrado, Hughes, and
  Dean}]{johnson-etal-2017-googles}
Melvin Johnson, Mike Schuster, Quoc~V. Le, Maxim Krikun, Yonghui Wu, Zhifeng
  Chen, Nikhil Thorat, Fernanda Vi{\'e}gas, Martin Wattenberg, Greg Corrado,
  Macduff Hughes, and Jeffrey Dean. 2017.
\newblock \href {https://doi.org/10.1162/tacl_a_00065} {{G}oogle{'}s
  multilingual neural machine translation system: Enabling zero-shot
  translation}.
\newblock \emph{Transactions of the Association for Computational Linguistics},
  5:339--351.

\bibitem[{Junczys-Dowmunt et~al.(2018)Junczys-Dowmunt, Grundkiewicz, Guha, and
  Heafield}]{junczys-dowmunt-etal-2018-approaching}
Marcin Junczys-Dowmunt, Roman Grundkiewicz, Shubha Guha, and Kenneth Heafield.
  2018.
\newblock \href {https://doi.org/10.18653/v1/N18-1055} {Approaching neural
  grammatical error correction as a low-resource machine translation task}.
\newblock In \emph{Proceedings of the 2018 Conference of the North {A}merican
  Chapter of the Association for Computational Linguistics: Human Language
  Technologies, Volume 1 (Long Papers)}, pages 595--606, New Orleans,
  Louisiana. Association for Computational Linguistics.

\bibitem[{Kaneko et~al.(2020)Kaneko, Mita, Kiyono, Suzuki, and
  Inui}]{kaneko-etal-2020-encoder}
Masahiro Kaneko, Masato Mita, Shun Kiyono, Jun Suzuki, and Kentaro Inui. 2020.
\newblock \href {https://doi.org/10.18653/v1/2020.acl-main.391}
  {Encoder-decoder models can benefit from pre-trained masked language models
  in grammatical error correction}.
\newblock In \emph{Proceedings of the 58th Annual Meeting of the Association
  for Computational Linguistics}, pages 4248--4254, Online. Association for
  Computational Linguistics.

\bibitem[{Kasewa et~al.(2018)Kasewa, Stenetorp, and
  Riedel}]{kasewa-etal-2018-wronging}
Sudhanshu Kasewa, Pontus Stenetorp, and Sebastian Riedel. 2018.
\newblock \href {https://doi.org/10.18653/v1/D18-1541} {Wronging a right:
  Generating better errors to improve grammatical error detection}.
\newblock In \emph{Proceedings of the 2018 Conference on Empirical Methods in
  Natural Language Processing}, pages 4977--4983, Brussels, Belgium.
  Association for Computational Linguistics.

\bibitem[{Kiyono et~al.(2019)Kiyono, Suzuki, Mita, Mizumoto, and
  Inui}]{gec-pseudo-data}
Shun Kiyono, Jun Suzuki, Masato Mita, Tomoya Mizumoto, and Kentaro Inui. 2019.
\newblock \href {https://doi.org/10.18653/v1/D19-1119} {An empirical study of
  incorporating pseudo data into grammatical error correction}.
\newblock In \emph{Proceedings of the 2019 Conference on Empirical Methods in
  Natural Language Processing and the 9th International Joint Conference on
  Natural Language Processing (EMNLP-IJCNLP)}, pages 1236--1242, Hong Kong,
  China. Association for Computational Linguistics.

\bibitem[{Lichtarge et~al.(2020)Lichtarge, Alberti, and Kumar}]{jared-tacl}
Jared Lichtarge, Chris Alberti, and Shankar Kumar. 2020.
\newblock \href {https://doi.org/10.1162/tacl_a_00336} {Data weighted training
  strategies for grammatical error correction}.
\newblock \emph{Transactions of the Association for Computational Linguistics},
  8:634--646.

\bibitem[{Lichtarge et~al.(2019)Lichtarge, Alberti, Kumar, Shazeer, Parmar, and
  Tong}]{lichtarge-etal-2019-corpora}
Jared Lichtarge, Chris Alberti, Shankar Kumar, Noam Shazeer, Niki Parmar, and
  Simon Tong. 2019.
\newblock \href {https://doi.org/10.18653/v1/N19-1333} {Corpora generation for
  grammatical error correction}.
\newblock In \emph{Proceedings of the 2019 Conference of the North {A}merican
  Chapter of the Association for Computational Linguistics: Human Language
  Technologies, Volume 1 (Long and Short Papers)}, pages 3291--3301,
  Minneapolis, Minnesota. Association for Computational Linguistics.

\bibitem[{Mizumoto et~al.(2011)Mizumoto, Komachi, Nagata, and
  Matsumoto}]{mizumoto-etal-2011-mining}
Tomoya Mizumoto, Mamoru Komachi, Masaaki Nagata, and Yuji Matsumoto. 2011.
\newblock \href {https://www.aclweb.org/anthology/I11-1017} {Mining revision
  log of language learning {SNS} for automated {J}apanese error correction of
  second language learners}.
\newblock In \emph{Proceedings of 5th International Joint Conference on Natural
  Language Processing}, pages 147--155, Chiang Mai, Thailand. Asian Federation
  of Natural Language Processing.

\bibitem[{Napoles et~al.(2015)Napoles, Sakaguchi, Post, and
  Tetreault}]{napoles-etal-2015-ground}
Courtney Napoles, Keisuke Sakaguchi, Matt Post, and Joel Tetreault. 2015.
\newblock \href {https://doi.org/10.3115/v1/P15-2097} {Ground truth for
  grammatical error correction metrics}.
\newblock In \emph{Proceedings of the 53rd Annual Meeting of the Association
  for Computational Linguistics and the 7th International Joint Conference on
  Natural Language Processing (Volume 2: Short Papers)}, pages 588--593,
  Beijing, China. Association for Computational Linguistics.

\bibitem[{Napoles et~al.(2017)Napoles, Sakaguchi, and
  Tetreault}]{napoles-etal-2017-jfleg}
Courtney Napoles, Keisuke Sakaguchi, and Joel Tetreault. 2017.
\newblock \href {https://www.aclweb.org/anthology/E17-2037} {{JFLEG}: A fluency
  corpus and benchmark for grammatical error correction}.
\newblock In \emph{Proceedings of the 15th Conference of the {E}uropean Chapter
  of the Association for Computational Linguistics: Volume 2, Short Papers},
  pages 229--234, Valencia, Spain. Association for Computational Linguistics.

\bibitem[{Ng et~al.(2013)Ng, Wu, Wu, Hadiwinoto, and
  Tetreault}]{ng-etal-2013-conll}
Hwee~Tou Ng, Siew~Mei Wu, Yuanbin Wu, Christian Hadiwinoto, and Joel Tetreault.
  2013.
\newblock \href {https://www.aclweb.org/anthology/W13-3601} {The {C}o{NLL}-2013
  shared task on grammatical error correction}.
\newblock In \emph{Proceedings of the Seventeenth Conference on Computational
  Natural Language Learning: Shared Task}, pages 1--12, Sofia, Bulgaria.
  Association for Computational Linguistics.

\bibitem[{Omelianchuk et~al.(2020)Omelianchuk, Atrasevych, Chernodub, and
  Skurzhanskyi}]{gector}
Kostiantyn Omelianchuk, Vitaliy Atrasevych, Artem Chernodub, and Oleksandr
  Skurzhanskyi. 2020.
\newblock \href {https://doi.org/10.18653/v1/2020.bea-1.16} {{GECT}o{R} {--}
  grammatical error correction: Tag, not rewrite}.
\newblock In \emph{Proceedings of the Fifteenth Workshop on Innovative Use of
  NLP for Building Educational Applications}, pages 163--170, Seattle, WA, USA
  Online. Association for Computational Linguistics.

\bibitem[{Raffel et~al.(2020)Raffel, Shazeer, Roberts, Lee, Narang, Matena,
  Zhou, Li, and Liu}]{c4}
Colin Raffel, Noam Shazeer, Adam Roberts, Katherine Lee, Sharan Narang, Michael
  Matena, Yanqi Zhou, Wei Li, and Peter~J. Liu. 2020.
\newblock \href {http://jmlr.org/papers/v21/20-074.html} {Exploring the limits
  of transfer learning with a unified text-to-text transformer}.
\newblock \emph{Journal of Machine Learning Research}, 21(140):1--67.

\bibitem[{Rei et~al.(2017)Rei, Felice, Yuan, and
  Briscoe}]{rei-etal-2017-artificial}
Marek Rei, Mariano Felice, Zheng Yuan, and Ted Briscoe. 2017.
\newblock \href {https://doi.org/10.18653/v1/W17-5032} {Artificial error
  generation with machine translation and syntactic patterns}.
\newblock In \emph{Proceedings of the 12th Workshop on Innovative Use of {NLP}
  for Building Educational Applications}, pages 287--292, Copenhagen, Denmark.
  Association for Computational Linguistics.

\bibitem[{Rozovskaya and Roth(2010)}]{rozovskaya-roth-2010-generating}
Alla Rozovskaya and Dan Roth. 2010.
\newblock \href {https://www.aclweb.org/anthology/D10-1094} {Generating
  confusion sets for context-sensitive error correction}.
\newblock In \emph{Proceedings of the 2010 Conference on Empirical Methods in
  Natural Language Processing}, pages 961--970, Cambridge, MA. Association for
  Computational Linguistics.

\bibitem[{Schrijver(2003)}]{schrijver-book}
A.~Schrijver. 2003.
\newblock \emph{Combinatorial Optimization - Polyhedra and Efficiency}.
\newblock Springer.

\bibitem[{Sennrich et~al.(2016)Sennrich, Haddow, and
  Birch}]{sennrich-etal-2016-improving}
Rico Sennrich, Barry Haddow, and Alexandra Birch. 2016.
\newblock \href {https://doi.org/10.18653/v1/P16-1009} {Improving neural
  machine translation models with monolingual data}.
\newblock In \emph{Proceedings of the 54th Annual Meeting of the Association
  for Computational Linguistics (Volume 1: Long Papers)}, pages 86--96, Berlin,
  Germany. Association for Computational Linguistics.

\bibitem[{Shazeer and Stern(2018)}]{adafactor}
Noam Shazeer and Mitchell Stern. 2018.
\newblock Adafactor: Adaptive learning rates with sublinear memory cost.
\newblock \emph{arXiv preprint arXiv:1804.04235}.

\bibitem[{Stahlberg and Byrne(2019)}]{stahlberg-byrne-2019-cueds}
Felix Stahlberg and Bill Byrne. 2019.
\newblock \href {https://doi.org/10.18653/v1/W19-4417} {The {CUED}{'}s
  grammatical error correction systems for {BEA}-2019}.
\newblock In \emph{Proceedings of the Fourteenth Workshop on Innovative Use of
  NLP for Building Educational Applications}, pages 168--175, Florence, Italy.
  Association for Computational Linguistics.

\bibitem[{Stahlberg and Kumar(2020)}]{seq2edits}
Felix Stahlberg and Shankar Kumar. 2020.
\newblock \href {https://doi.org/10.18653/v1/2020.emnlp-main.418}
  {{S}eq2{E}dits: Sequence transduction using span-level edit operations}.
\newblock In \emph{Proceedings of the 2020 Conference on Empirical Methods in
  Natural Language Processing (EMNLP)}, pages 5147--5159, Online. Association
  for Computational Linguistics.

\bibitem[{Takahashi et~al.(2020)Takahashi, Katsumata, and
  Komachi}]{takahashi-etal-2020-grammatical}
Yujin Takahashi, Satoru Katsumata, and Mamoru Komachi. 2020.
\newblock \href {https://doi.org/10.18653/v1/2020.acl-srw.5} {Grammatical error
  correction using pseudo learner corpus considering learner{'}s error
  tendency}.
\newblock In \emph{Proceedings of the 58th Annual Meeting of the Association
  for Computational Linguistics: Student Research Workshop}, pages 27--32,
  Online. Association for Computational Linguistics.

\bibitem[{Vaswani et~al.(2018)Vaswani, Bengio, Brevdo, Chollet, Gomez, Gouws,
  Jones, Kaiser, Kalchbrenner, Parmar, Sepassi, Shazeer, and
  Uszkoreit}]{vaswani-etal-2018-tensor2tensor}
Ashish Vaswani, Samy Bengio, Eugene Brevdo, Francois Chollet, Aidan Gomez,
  Stephan Gouws, Llion Jones, {\L}ukasz Kaiser, Nal Kalchbrenner, Niki Parmar,
  Ryan Sepassi, Noam Shazeer, and Jakob Uszkoreit. 2018.
\newblock \href {https://www.aclweb.org/anthology/W18-1819} {{T}ensor2{T}ensor
  for neural machine translation}.
\newblock In \emph{Proceedings of the 13th Conference of the Association for
  Machine Translation in the {A}mericas (Volume 1: Research Papers)}, pages
  193--199, Boston, MA. Association for Machine Translation in the Americas.

\bibitem[{Vaswani et~al.(2017)Vaswani, Shazeer, Parmar, Uszkoreit, Jones,
  Gomez, Kaiser, and Polosukhin}]{transformer}
Ashish Vaswani, Noam Shazeer, Niki Parmar, Jakob Uszkoreit, Llion Jones,
  Aidan~N Gomez, \L~ukasz Kaiser, and Illia Polosukhin. 2017.
\newblock \href
  {http://papers.nips.cc/paper/7181-attention-is-all-you-need.pdf} {Attention
  is all you need}.
\newblock In I.~Guyon, U.~V. Luxburg, S.~Bengio, H.~Wallach, R.~Fergus,
  S.~Vishwanathan, and R.~Garnett, editors, \emph{Advances in Neural
  Information Processing Systems 30}, pages 5998--6008. Curran Associates, Inc.

\bibitem[{Wan et~al.(2020)Wan, Wan, and Wang}]{wan-etal-2020-improving}
Zhaohong Wan, Xiaojun Wan, and Wenguang Wang. 2020.
\newblock \href {https://doi.org/10.18653/v1/2020.coling-main.200} {Improving
  grammatical error correction with data augmentation by editing latent
  representation}.
\newblock In \emph{Proceedings of the 28th International Conference on
  Computational Linguistics}, pages 2202--2212, Barcelona, Spain (Online).
  International Committee on Computational Linguistics.

\bibitem[{Xie et~al.(2018)Xie, Genthial, Xie, Ng, and
  Jurafsky}]{xie-etal-2018-noising}
Ziang Xie, Guillaume Genthial, Stanley Xie, Andrew Ng, and Dan Jurafsky. 2018.
\newblock \href {https://doi.org/10.18653/v1/N18-1057} {Noising and denoising
  natural language: Diverse backtranslation for grammar correction}.
\newblock In \emph{Proceedings of the 2018 Conference of the North {A}merican
  Chapter of the Association for Computational Linguistics: Human Language
  Technologies, Volume 1 (Long Papers)}, pages 619--628, New Orleans,
  Louisiana. Association for Computational Linguistics.

\bibitem[{Xu et~al.(2019)Xu, Zhang, Chen, and Qin}]{xu-etal-2019-erroneous}
Shuyao Xu, Jiehao Zhang, Jin Chen, and Long Qin. 2019.
\newblock \href {https://doi.org/10.18653/v1/W19-4415} {Erroneous data
  generation for grammatical error correction}.
\newblock In \emph{Proceedings of the Fourteenth Workshop on Innovative Use of
  NLP for Building Educational Applications}, pages 149--158, Florence, Italy.
  Association for Computational Linguistics.

\bibitem[{Yannakoudakis et~al.(2011)Yannakoudakis, Briscoe, and
  Medlock}]{yannakoudakis-etal-2011-new}
Helen Yannakoudakis, Ted Briscoe, and Ben Medlock. 2011.
\newblock \href {https://www.aclweb.org/anthology/P11-1019} {A new dataset and
  method for automatically grading {ESOL} texts}.
\newblock In \emph{Proceedings of the 49th Annual Meeting of the Association
  for Computational Linguistics: Human Language Technologies}, pages 180--189,
  Portland, Oregon, USA. Association for Computational Linguistics.

\bibitem[{Zhao et~al.(2019)Zhao, Wang, Shen, Jia, and
  Liu}]{zhao-etal-2019-improving}
Wei Zhao, Liang Wang, Kewei Shen, Ruoyu Jia, and Jingming Liu. 2019.
\newblock \href {https://doi.org/10.18653/v1/N19-1014} {Improving grammatical
  error correction via pre-training a copy-augmented architecture with
  unlabeled data}.
\newblock In \emph{Proceedings of the 2019 Conference of the North {A}merican
  Chapter of the Association for Computational Linguistics: Human Language
  Technologies, Volume 1 (Long and Short Papers)}, pages 156--165, Minneapolis,
  Minnesota. Association for Computational Linguistics.

\end{thebibliography}
\bibliographystyle{acl_natbib}

\clearpage

\appendix

\section{The Seq2Edits Model}
\label{sec:seq2edits}

\begin{figure*}[t!]
\centering
\small
\includegraphics[scale=0.35]{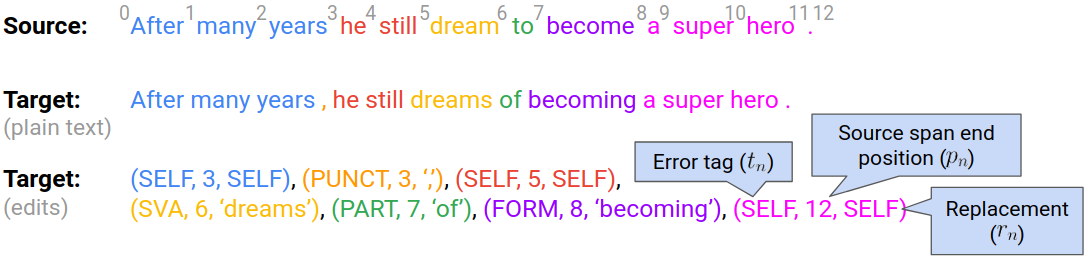}
\caption{Representing grammatical error correction as a sequence of span-based edit operations. The implicit start position for a source span is the end position of the previous edit operation. \texttt{SELF} indicates spans that are copied over from the source sentence ($\mathbf{x}$). The probability of the first two edits is given by:
$P(\text{After many years ,}|\mathbf{x}) =
{\color{lightblue} P(t_1=\text{SELF}|\mathbf{x}) \cdot P(p_1=3|\text{SELF},\mathbf{x}) \cdot P(r_1=\text{SELF}|\text{SELF},3,\mathbf{x})} \cdot
{\color{darkyellow} P(t_2=\text{PUNCT}|\text{SELF},3,\text{SELF},\mathbf{x}) \cdot P(p_2=3|\text{SELF},3,\text{SELF},\text{PUNCT},\mathbf{x})
\cdot P(r_2=\text{,}|\text{SELF},3,\text{SELF},\text{PUNCT},3,\mathbf{x})}
$. Figure and caption are taken from \citet{seq2edits}.}
\label{fig:edit-ops}
\end{figure*}

\begin{figure*}[t!]
\centering
\small
\includegraphics[scale=0.9]{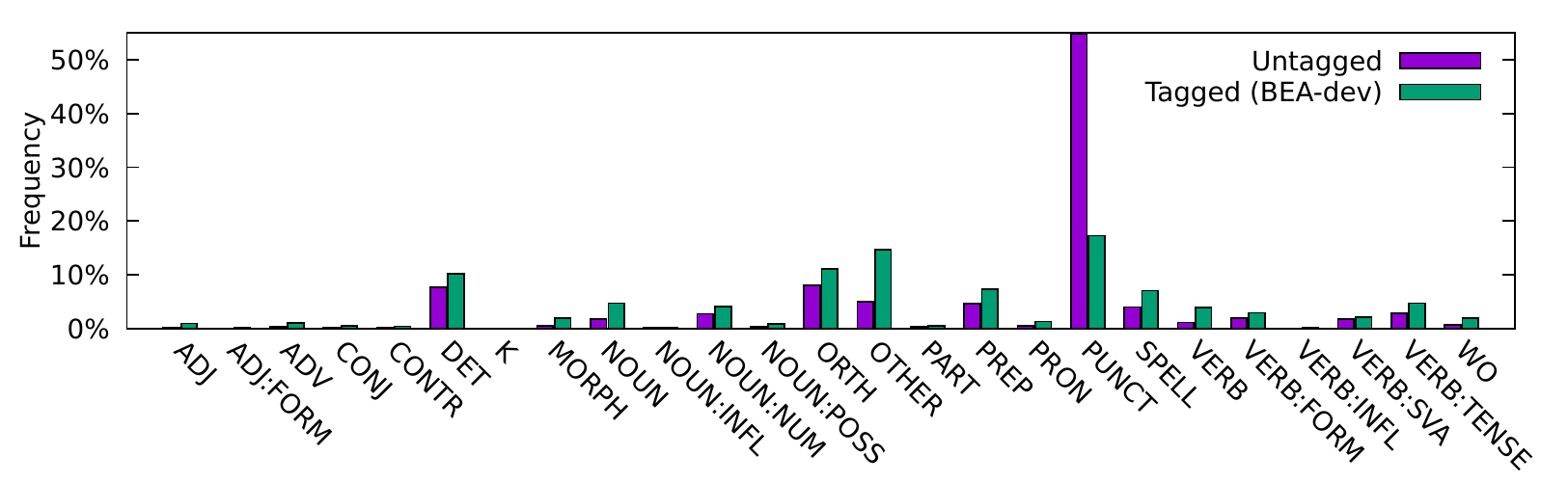}
\caption{Error tag distributions for untagged and tagged Seq2Edits corruption models. Most of the untagged corruptions are dull punctuation errors (\texttt{PUNCT}) whereas the tagged corruptions are more diverse as they follow the BEA-dev tag distribution in Fig.~\ref{fig:distributions}.}
\label{fig:distributions-corrupt}
\end{figure*}

This section contains a short description of the key elements of the Seq2Edits model which we use extensively in this work. For a detailed discussion we refer the reader to the full paper from \citet{seq2edits}.

Seq2Edits represents sequence transduction as a sequence of edit operations. Each edit operation is applied to a span in the source sentence which is either copied (labelled with \texttt{SELF}) or replaced by replacement tokens. Fig.\ \ref{fig:edit-ops} shows that each edit is represented by a 3-tuple consisting of an error tag $t_n$, the source span end position $p_n$ (the implicit source span start position is $p_{n-1}$), and the replacement token $r_n$. The sequence of 3-tuples is predicted auto-regressively by a modified Transformer with multiple target tapes.

Unlike \citet{seq2edits}, in this work we use Seq2Edits as a corruption model. Thus, each edit corresponds to an artificial error of a certain type (represented by the error tag $t_n$) rather than a correction. In Sec.\ \ref{sec:tagged-corruption} we constrain the error tag tape with FSTs (Fig.\ \ref{fig:constraint-fsts}) to force the generation of a specific error type.

\section{Example Corruptions}
\label{sec:example-corruptions}

The main contribution of this work is using explicit error type tags to control and diversify the output of corruption models since conventional corruption models without tags tend to generate dull and monotonous output. In the examples in Table \ref{tab:examples-ceg} the untagged corruption model simply deletes a determiner while the tagged corruption model is able to produce a wide variety of human-like writing errors. This observation is supported by Fig.~\ref{fig:distributions-corrupt} which shows that the untagged corruption model mainly outputs simplistic punctuation errors while the untagged model has a much better coverage of other more complex error types.

\begin{table*}[t!]
\centering
\small
\begin{tabularx}{\linewidth}{lX}
\hline
Input (clean) & I'm learning a lot and the students are very friendly. \\
Untagged corruption (1-best) & I'm learning a lot and the students are very friendly. \\ 
Untagged corruption (2-best) & I'm learning a lot \textbf{and students} are very friendly. \\ 
Tagged corruption & \\
\ \ \ \texttt{ADJ} & I'm learning a lot and the students are very \textbf{friendliness}. \\
\ \ \ \texttt{ADJ:FORM} & I'm learning a lot and the students are very \textbf{friendlies}. \\
\ \ \ \texttt{ADV} & I'm learning a lot and the students are \textbf{so} friendly. \\
\ \ \ \texttt{CONJ} & I'm learning a \textbf{lot the} students are very friendly. \\
\ \ \ \texttt{CONTR} & \textbf{I learning} a lot and the students are very friendly. \\
\ \ \ \texttt{DET} & I'm learning a lot \textbf{and students} are very friendly. \\
\ \ \ \texttt{K} & \textbf{I 'm} learning a lot and the students are very friendly. \\
\ \ \ \texttt{MORPH} & I'm learning a lot and the students are very \textbf{friendship}. \\
\ \ \ \texttt{NOUN} & I'm learning \textbf{many things} and the students are very friendly. \\
\ \ \ \texttt{NOUN:INFL} & I'm learning a lot and the \textbf{studentes} are very friendly. \\
\ \ \ \texttt{NOUN:NUM} & I'm learning a lot and the \textbf{student are} very friendly. \\
\ \ \ \texttt{NOUN:POSS} & I'm learning a lot and the \textbf{student's} are very friendly. \\
\ \ \ \texttt{ORTH} & I'm learning \textbf{alot} and the students are very friendly. \\
\ \ \ \texttt{OTHER} & I'm learning \textbf{very much} and the students are very friendly. \\
\ \ \ \texttt{PART} & I'm learning \textbf{up} a lot and the students are very friendly. \\
\ \ \ \texttt{PREP} & I'm learning \textbf{to} a lot and the students are very friendly. \\
\ \ \ \texttt{PRON} & \textbf{Learning} a lot and the students are very friendly. \\
\ \ \ \texttt{PUNCT} & I'm learning a lot and the students are very \textbf{friendly} \\
\ \ \ \texttt{SPELL} & I'm \textbf{lerning} a lot and the students are very friendly. \\
\ \ \ \texttt{VERB} & I'm learning a lot and the \textbf{students very} friendly. \\
\ \ \ \texttt{VERB:FORM} & I'm \textbf{learn} a lot and the students are very friendly. \\
\ \ \ \texttt{VERB:INFL} & I'm \textbf{learnes} a lot and the students are very friendly. \\
\ \ \ \texttt{VERB:SVA} & I'm learning a lot and the \textbf{students is} very friendly. \\
\ \ \ \texttt{VERB:TENSE} & \textbf{I learn} a lot and the students are very friendly. \\
\ \ \ \texttt{WO} & \textbf{I'm a lot learning} and the students are very friendly. \\
\hline
Input (clean) & The British summertime was first introduced in England in 1908. \\
Untagged corruption (1-best) & The British summertime was first introduced in England in 1908. \\ 
Untagged corruption (2-best) & \textbf{British} summertime was first introduced in England in 1908. \\ 
Tagged corruption & \\
\ \ \ \texttt{ADJ} & The \textbf{English} summertime was first introduced in England in 1908. \\
\ \ \ \texttt{ADJ:FORM} &The \textbf{Britishest} summertime was first introduced in England in 1908. \\
\ \ \ \texttt{ADV} & The British summertime \textbf{was introduced} in England in 1908. \\
\ \ \ \texttt{CONJ} & \textbf{And British} summertime was first introduced in England in 1908.\\
\ \ \ \texttt{CONTR} & The British \textbf{summertime's} first introduced in England in 1908. \\
\ \ \ \texttt{DET} & \textbf{British} summertime was first introduced in England in 1908. \\
\ \ \ \texttt{K} & British summertime was introduced in \textbf{England-in} 1908. \\
\ \ \ \texttt{MORPH} & The \textbf{Britishes} summertime was first introduced in England in 1908. \\
\ \ \ \texttt{NOUN} & The British \textbf{summer} was first introduced in England in 1908. \\
\ \ \ \texttt{NOUN:INFL} & The British \textbf{Summertimes} was first introduced in England in 1908. \\
\ \ \ \texttt{NOUN:NUM} & The British \textbf{Summertimes} was first introduced in England in 1908. \\
\ \ \ \texttt{NOUN:POSS} & The \textbf{British's} summertime was first introduced in England in 1908. \\
\ \ \ \texttt{ORTH} &  The British \textbf{summer time} was first introduced in England in 1908.\\
\ \ \ \texttt{OTHER} & The British summertime was introduced \textbf{for first time} in England in 1908. \\
\ \ \ \texttt{PART} & The British summertime was first introduced \textbf{to} in England in 1908. \\
\ \ \ \texttt{PREP} & The British summertime was first introduced \textbf{to} England in 1908. \\
\ \ \ \texttt{PRON} &  \textbf{It's} British summertime was first introduced in England in 1908. \\
\ \ \ \texttt{PUNCT} & The British summer time was first introduced in England in \textbf{1908} \\
\ \ \ \texttt{SPELL} & The \textbf{Britishi} summertime was first introduced in England in 1908. \\
\ \ \ \texttt{VERB} & British summertime was first \textbf{invented} in England in 1908. \\
\ \ \ \texttt{VERB:FORM} & The British summertime was first \textbf{introduce} in England in 1908. \\
\ \ \ \texttt{VERB:INFL} & The British summertime was first \textbf{introduceed} in England in 1908. \\
\ \ \ \texttt{VERB:SVA} & The British summertime \textbf{were} first introduced in England in 1908. \\
\ \ \ \texttt{VERB:TENSE} & The British summertime \textbf{is} first introduced in England in 1908. \\
\ \ \ \texttt{WO} & The British summertime was \textbf{introduced first} in England in 1908. \\
\hline
\end{tabularx}
\caption{\label{tab:examples-ceg} Example outputs from tagged and untagged Seq2Edits corruption models.}
\end{table*}

\section{Example Corrections}
\label{sec:example-outputs}

\begin{table*}[b!]
\centering
\small
\begin{tabularx}{\linewidth}{lX}
\hline
Source & In my \textbf{country, taipei, we always stock in traffic} for about one hour in the morning. \\
Reference & In my \textbf{city, Taipei, we are always stuck in traffic} for about one hour in the morning. \\ 
Output (without C4$_\text{200M}$) & In my \textbf{country, Taipei, we always stock up on traffic} for about one hour in the morning. \\
Output (with C4$_\text{200M}$) & In my \textbf{country, Taipei, we are always stuck in traffic} for about one hour in the morning. \\
\hline
Source & Back to the topic, I don't know \textbf{well} any of member of in my family\textbf{,} I don't trust them, \textbf{honestly}. \\
Reference & Back to the topic. I don't know any of the members of my family \textbf{well.} I don't trust them, \textbf{to be honest}. \\ 
Output (without C4$_\text{200M}$) & Back to the topic, I don't know \textbf{well} any members of my family\textbf{,} I don't trust them, \textbf{honestly}. \\
Output (with C4$_\text{200M}$) & Back to the topic, I don't know any members of my family \textbf{well,} I don't trust them, \textbf{honestly}. \\
\hline
Source & What \textbf{is the ``Family'' meaning?} \\
Reference & What \textbf{does ``Family'' mean?} \\ 
Output (without C4$_\text{200M}$) & What \textbf{is the ``Family'' meaning?} \\
Output (with C4$_\text{200M}$) & What \textbf{does ``Family'' mean?} \\
\hline
Source & Basketball has many benefits \textbf{not to me but also everyone.} \\
Reference & Basketball has many benefits\textbf{, not just for me but also for everyone.} \\ 
Output (without C4$_\text{200M}$) & Basketball has many benefits \textbf{not to me but also everyone.} \\
Output (with C4$_\text{200M}$) & Basketball has many benefits \textbf{not only for me but also for everyone.} \\
\hline
Source & The manager of \textbf{all this} project is Miss June Sid, our science teacher. \\
Reference & The manager of \textbf{this} project is Miss June Sid, our science teacher. \\ 
Output (without C4$_\text{200M}$) & The manager of \textbf{all this} project is Miss June Sid, our science teacher. \\
Output (with C4$_\text{200M}$) & The manager of \textbf{this} project is Miss June Sid, our science teacher. \\
\hline
Source & Volleyball is a sport \textbf{play every place, when} I \textbf{travel} on the beach I like \textbf{plays} with my sister in the sand and \textbf{after} we \textbf{are going to} the sea. \\
Reference & Volleyball is a sport \textbf{that is played everywhere. When} I \textbf{am} on the beach I like \textbf{playing} with my sister in the sand and \textbf{then} we \textbf{go in} the sea. \\ 
Output (without C4$_\text{200M}$) &  Volleyball is a sport \textbf{played in every place. When} I \textbf{travel} on the beach I like \textbf{to play} with my sister in the sand and \textbf{after} we \textbf{go to} the sea. \\
Output (with C4$_\text{200M}$) &  Volleyball is a sport \textbf{played everywhere. When} I \textbf{travel} on the beach, I like \textbf{playing} with my sister in the sand and \textbf{after} we \textbf{go to} the sea. \\
\hline
Source & Because public transport is \textbf{a cost effective} and better resource allocation in mass transport system. \\
Reference & Because public transport is \textbf{a more cost-effective} and better resource allocation in mass transport system. \\ 
Output (without C4$_\text{200M}$) & Because public transport is \textbf{a cost effective} and better resource allocation in \textbf{a} mass transport system. \\
Output (with C4$_\text{200M}$) & Because public transport is \textbf{cost effective} and \textbf{there} is better resource allocation in \textbf{the} mass transport system. \\
\hline
\end{tabularx}
\caption{\label{tab:examples-gec} Example outputs on BEA-dev~\citep{bryant-etal-2019-bea} after two stage fine-tuning. The system without C4$_\text{200M}$ corresponds to the first row in Fig.\ \ref{fig:pre-bar-chart}. The outputs with C4$_\text{200M}$ were generated from the system in Table~\ref{tab:cross-dev-set}e.}
\end{table*}

Table \ref{tab:examples-gec} shows example outputs from two systems: with and without using our new C4$_\text{200M}$ pre-training data set.\footnote{\href{https://github.com/google-research-datasets/C4_200M-synthetic-dataset-for-grammatical-error-correction}{https://github.com/google-research-datasets/C4\_200M-synthetic-dataset-for-grammatical-error-correction}} The system with C4$_\text{200M}$ tends to be more fluent because it is trained on much more English text from diverse sources. The first example in Table~\ref{tab:examples-gec} demonstrates that pre-training on C4$_\text{200M}$ also seems to help learning semantics -- the non-C4$_\text{200M}$ output (``we always stock up on traffic'') is grammatically correct, but the C4$_\text{200M}$ model has learned that ``stocking up on traffic'' is nonsensical, and that being ``stuck in traffic'' is probably a better match for the author's intent.

The system with C4$_\text{200M}$ is able to work well across a wider range of domains by proposing more radical changes such as long-range reorderings to improve the fluency as demonstrated by the movement of ``well'' in the second example or the major rewrite in the third example of Table \ref{tab:examples-gec}. However, this can make the C4$_\text{200M}$ model more prone to subtle changes in semantics, as shown in the last example (``public transport is a cost effective (..) allocation'' $\rightarrow$ ``public transport is cost effective and ..'').

\end{document}